\documentclass[a4paper,twoside]{article}

\usepackage{epsfig}
\usepackage{subcaption}
\usepackage{calc}
\usepackage{amssymb}
\usepackage{amstext}
\usepackage{amsmath}
\usepackage{amsthm}
\usepackage{multicol}
\usepackage{multirow}

\usepackage{color}
\usepackage{epstopdf}
\usepackage{graphicx}
\usepackage{xcolor}
\usepackage{amsmath}
\usepackage{multirow}

\usepackage{ifthen} 
\usepackage{graphicx}
\usepackage{subcaption}
\usepackage{verbatim}
\usepackage{xfrac}

\usepackage{SCITEPRESS}     % Please add other packages that you may need BEFORE the SCITEPRESS.sty package.
\usepackage{natbib}

\newcommand{\x}{\text{\bf x}}
\newcommand{\y}{\text{\bf y}}

\definecolor{dg}{rgb}{0.0, 0.0, 0.0}
\definecolor{or}{rgb}{0.0, 0.0, 0.0}
\definecolor{mei}{rgb}{0.0, 0.0, 0.0}

\newcommand{\dimi}[1]{\textcolor{dg}{#1}}
\newboolean{anonymizeIt} 
\setboolean{anonymizeIt}{false} 
\usepackage[german,textsize=footnotesize,textwidth=1.6cm]{todonotes}
\usepackage{marginnote}

\begin{document}
	
	\title{Polygonizing roof segments from high-resolution aerial images using YOLOv8-based edge detection}
	\ifthenelse{\boolean{anonymizeIt}}{

\author{
			\authorname{None}
            \affiliation{None}}
 
}{
		\author{
			\authorname{Qipeng Mei\sup{1}\orcidAuthor{0009-0006-0506-4478}, Dimitri Bulatov\sup{2}\orcidAuthor{0000-0002-0560-2591}, Dorota Iwaszczuk\sup{1}\orcidAuthor{0000-0002-5969-8533}}
            \affiliation{\sup{1}\scriptsize{Remote Sensing and Image Analysis, Department of Civil and Environmental Engineering Sciences, Technical University of Darmstadt, Franziska-Braun-Str. 7, 64287 Darmstadt, Germany}}
			\affiliation{\sup{2}\scriptsize{Fraunhofer IOSB Ettlingen, Gutleuthausstrasse 1, 76275 Ettlingen, Germany}}}
		
	%\email{\{dimitri.bulatov@iosb.fraunhofer.de\}}	
	}
	%The paper must have at least one keyword. The text must be set to 9-point font size and without the use of bold or italic font style. For more than one keyword, please use a comma as a separator. Keywords must be titlecased.}
	
	\abstract{
		This study presents a novel approach for roof detail extraction and vectorization using remote sensing images. 
  Unlike previous geometric-primitive-based methods that rely on the detection of corners, our method focuses on edge detection as the primary mechanism for roof reconstruction, while utilizing geometric relationships to define corners and faces.
  We adapt the YOLOv8 OBB model, originally designed for rotated object detection, to extract roof edges effectively.
  Our method demonstrates robustness against noise and occlusion, leading to precise vectorized representations of building roofs.
  Experiments conducted on the SGA and Melville datasets highlight the method’s effectiveness.
  At the raster level, our model outperforms the state-of-the-art foundation segmentation model (SAM), achieving a mIoU between 0.85 and 1 for most samples and an ovIoU close to 0.97. 
  At the vector level, evaluation using the Hausdorff distance, PolyS metric, and our raster-vector-metric demonstrates significant improvements after polygonization, with a close approximation to the reference data.
  The method successfully handles diverse roof structures and refines edge gaps, even on complex roof structures of new, excluded from training datasets.
  Our findings underscore the potential of this approach to address challenges in automatic roof structure vectorization, supporting various applications such as urban terrain reconstruction. 
	}
	\onecolumn \maketitle \normalsize \setcounter{footnote}{0} \vfill

\section{MOTIVATION}\label{sec:motivation}

From traditional Chinese pavilions to Mexican pyramids and from skylines of US-megapoli to the residential areas of Australian urbanities, building roofs are fascinating products of human creativity. 
Humans are motivated to be creative because a roof is the culmination of what some people call ``home'' and what for other people is the workplace and gives them self-realization, or it belongs to a public place, which must be presentable while looking at it from afar. 
By contemplating a building roof, we can easily discern its constituting elements, detect important corners and edges defining its structure, and infer the relations between these components. 
This ability to perceive structures from images is the fundamental aspect of human vision. 
However, retrieving very complex topological structures still presents a significant challenge \citep{lin2024optimized} for computer vision algorithms. 
This challenge negatively affects the ability to create automatically digital representations of large building databases from aerial images, posing an obstacle to those numerous applications requiring detailed knowledge of building roof geometry. 
The applications may require 3D information and the required output, consistently, would be the 3D geometry of roofs. 
Examples are environmental science,  planning mobile communication networks, real estate marketing, and virtual tourism, but also quick response applications \citep{bulatov2014context}. 
Alternatively, there could be purely 2D applications, which only require images: Roof panels for solar modules \citep{house2018using}, damage grading after some natural disasters \citep{lucks2019superpixel}, and, to a certain degree, urban planning. 
\par
These applications are particularly interesting for us because aerial images in high resolution are easily available nowadays whereas manual digitalization of building roof segments is costly. 
We therefore noticed a huge progress in this field made by scholars in the very few recent years. 
In what follows, a thorough literature research is carried out to identify the most promising tools but also certain shortcomings of related work that allowed us to propose a simple but efficient method to process buildings with complex roof graphs.

\subsection{Related Work}\label{sec:relatedwork}

A concept of \textbf{combining color region attribu\-tes and grouping} them to retrieve building roof details from images has been performed by \citet{henricsson1998role} mode than a quarter-century ago. 
However, irrelevant objects or disturbances, solar panels, chimneys, and shadows, with spectral characteristics other than the main roof negatively affect the computation of the region attributes. 
At least until nowadays, when the advanced foundational models, like SAM \citep{kirillov2023segment} appeared, this was the reason why the brilliant Henricsson's research of that time has not prevailed. 
Instead, in the absence of 3D data, scholars mostly concentrated on \textbf{building outlining}, a task successfully performed using conventional \citep{ZHANG1999,TURKER2015} and deep-learning based approaches \citep{wei2019toward,zorzi2022polyworld}. 
Unlike the vector representation of building outlines, which primarily deals with external contours, roof vectorization requires consideration of more complex topological structures, presenting a significant challenge.
To address this issue, the prevailing approach in current research moved, as already said, from color-based approaches towards retrieving \textbf{geometric primitives and} reconstructing the roof through \textbf{their topological relationships}.
Traditional geometric primitive detection methods, such as Harris corner detection \citep{harris1988combined} and Canny edge detection \citep{canny1986computational}, \dimi{are simple and robust, but}
% rely on image gradients. 
%While these techniques are able to effectively identify prominent features corresponding to roof boundaries in grayscale images, they have their limitations. 
the complexity of surface scenes and the impact of noise can result in the detection of geometric primitives associated with non-roof features as well as the omission or misidentification of roof-related primitives.
\par
\dimi{To grasp structures beyond local image gradient information}, the researchers have turned to deep learning approaches.
Probably, \citet{nauata2020vectorizing} were the first who used Convolutional Neural Networks (CNNs) to detect geometric primitives in aerial images. 
By combining these detections with integer programming, they inferred the relationships between the primitives and thus were able to assemble them into a cohesive planar graph.
In a similar vein, \citet{zhang2020conv} employed the Convolutional Message Passing Network (Conv-MPN) architecture to reconstruct roof structures. 
Basically, there are two networks, one for junction detection and one for establishing adjacency relationships. 
The method of \citet{hensel2021building} relies on the PPGNet Deep Neural network \citep{zhang2019ppgnet}, which is end-to-end trainable, and comprises modules for junction detection, line segment assignment, and adjacency matrix inference. 
The authors assess the weight for different losses and evaluate their results edge-wise. 
After the so-called DSM refinement step, supposed to detect, essentially, buildings and suppress vegetation, \citet{wang2021machine} uses a similar style transfer technique to detect rasterized roof corners and edges. 
These primitives are further refined,  slightly regularized, and employed to detect roof faces using a graph search algorithm. 
For each building, an undirected graph is built from the obtained edges.
A graph cycle, which is a roof polygon or a union of such, can be detected using a depth-first search. 
After all the cycles have been detected, large cycles that cover small cycles are removed to avoid face overlapping. 
This may work well for simple buildings, however, a choice of a minimum cycle basis is an exponential problem, and there is a need do combine geometrical and graph-theoretical considerations to solve this problem for complex buildings in an efficient way.  
\par
Another \textbf{remote-sensing-inspired} algorithm described in \citet{alidoost2020shaped} presupposes the application of a Y-shaped CNN from a single aerial image. The two outputs, which give the network its name, are style-transferred DSMs on the one hand, and the union of eave, ridge, and hip lines, on the other hand.   
Individual roof areas are retrieved using a post-processing step. 
This method is a further development of the knowledge-based workflow \citep{alidoost20192d}. 
Moreover, \citet{partovi2019automatic} proposed a comprehensive workflow consisting of building detection, decomposing of the roof into rectangles in 2D, ML-based assignment of every rectangle, as well as reconstruction of the roof because, for every rectangle, the set of parameters has to be determined. 
All possible models are instantiated by changing the parameters in the predefined ranges and validated by the PolyS metric \citep{avbelj2014metric}. 
%The authors denote their method hybrid, which means there are data-driven and model-driven modules in their workflow. 
%It could be argued that while using deep learning relying on large amounts of training examples, this differentiation has become obsolete because the models are provided implicitly, thanks to numerous examples. 
Even though for the assignment of roof types, the 3D information is dispensable, which makes this work very interesting for us, the dataset considered in this work shows typical neo-classic style buildings, extending in rectangular rows along the boulevards and designed in a uniform style. 
However, we are more interested in residential buildings since they have very complex roof models and are hardly decomposable into rectangles. 

\dimi{From the point of view of entire wire-frame generation,} \citet{zhao2022extracting} enhanced the detection units' ability to perceive line segment primitives by incorporating HT-IHT (Hough Transform-Iterative Hough Transform), which enabled the extraction of line segment and intersection point proposals. 
Subsequently, they used Graph Neural Networks (GNNs) to learn the relationships between line segments and intersections, thereby achieving the vectorization of roof structures. 
Inspired by the approach of \citet{huang2018learning}, \citet{esmaeily2023building} developed an end-to-end representation for intersection points in images. 
This representation not only captures the spatial coordinates of the intersection points but also encodes the direction of the line segments that form these intersections. 
Assisted by line segment detection masks and plane masks, this approach enables the vectorized representation of roof structures. 
\dimi{Wireframe extraction methods are efficient regarding computational resources, but sometimes unusual angles in challenging building structures may affect their performance negatively.}
%\dimi{???Qipeng, what was the problem about Esmaeily that we have identified? Let's finish this section with something negative :( so we can put your work into focus :). Besides, long time ago, you gave me this paper \cite{cao2023wireframenet}, is there something we can use on wireframe stuff from there??? }
\par
The work of \citet{ren2021intuitive}, despite mainly focusing on interactive roof annotation and roof graph optimization, also proposed a two-step Transformer and a GNN-based procedure. 
The transformer is supposed to retrieve a roof outline while the GNN is trained to predict the face adjacency. 
The approach even offers a tool to generate 3D models from 2D graphs using the so-called planarity metric, even though, as default, all inner roof vertices must have the same elevation. 
%Every edge of the outline and also every putative adjacency has a feature set. 
We exploit this contribution by retrieving their first dataset to validate our methods.
\par 
Finally, there are two very new and very successful methods worth mentioning. 
To detect geometric primitives and their relations, \citet{lin2024optimized} present a very special CNN denoted as Switch and emphasizing inter-channel rather than intra-channel (i.e. textural) patterns. 
The Transformer-based architecture called Roof-Former has been proposed by \citet{zhao2024vectorizing}. 
It consists of three steps: Feature Pyramid Networks providing relevant features encoding edges and vertices, Image feature fusion with enhanced  segmentation refinement, in which relevant feature sets seem to compete against each other, and Structural Reasoning. 
Qualitative results look impressive in the cases of misleadingly textured, blurred, etc. roofs.
The quantitative results are, however, given on the heat-map-level only. 

\subsection{Lessons Taken and Own Contributions}
In the existing work, the first clear message is that roof junction points are the most important geometric primitives in achieving vectorization. 
In other words, the lack or error in intersection point detection can severely degrade the results of roof structure vectorization. 
However, during the acquisition process, intersection points are susceptible to occlusion and noise, which is an objective reality that even the most advanced detection units cannot avoid. 
Therefore, our approach focuses on another important geometric primitive, namely edges. 
Edge detection is more stable and plays a role in connecting various elements in geometric structures. 
In our approach, we use a rotational object detection model (YOLOv8 OBB) to infer the vector representation of edges. 
\par
\dimi{The second important conclusion is that it is still state of the art to use some conventional post-processing because  a deep learning method cannot cover all possible roof structures and perceive all the roof details the architects are able to fantasize. 
The additional advantage of such a post-processing step is that rich findings from 3D-based roof detail analysis workflows, where the edges are intersection lines of some RANSAC-retrieved planes, are available and can be adopted \citep{verma20063d,SJJK,pohl2015gap,meidow2016enhancement,jung2017implicit} and many others.}
% VE,rottensteiner2010roof,bulatov2011context,bulatov2014context,
%From the geometric relationships, 
To the line end-points, we apply the Density-Based
Spatial Clustering of Applications with Noise (DBSCAN) algorithm \citep{sander1998density} which helps to gradually recover the intersection points and plane information, thereby achieving the vectorization of roof structures. 
\textcolor{mei}{\cite{bulatov2017chain} utilized DBSCAN to simplify the vector structure of the road network and to recognize junctions or dead ends. 
Inspired by them, in our case, it is used to cluster the endpoints of edge based on their spatial proximity to identify the potential junction points.}
\par
We apply our method to two challenging datasets and present both raster- and vector-based evaluation metrics. 
In particular, the PolyS metric \citep{avbelj2014metric}, widely used in remote sensing, has been applied for the first time, to our knowledge to 2D building roof polygons reconstructed using only images and no 3D data. Summarizing, 

\begin{itemize}
    \item We applied the YOLOv8 OBB method for detection of roof edges in high-resolution nadir aerial images;
    \item we developed a procedure for roof polygonization and face retrieval that relies on topology \emph{and} geometry, allowing  processing complex buildings;
    \item we evaluated our results on two datasets, one of which was not used for training but to track the model's ability to generalize. Hereby, we used both raster- and vector-based metrics, including the PolyS metric and our vector-raster-quality metric.
    \item  As a competing approach, we use (naive) SAM to evaluate its capability to extract roof faces.
\end{itemize}

\section{METHODOLOGY}\label{sec:methodology}

YOLOv8, the latest generation of the You Only Look Once (YOLO) model developed by \citet{yolov8_ultralytics}, extends its capabilities across a broad spectrum of computer vision tasks, including object detection, instance segmentation, pose estimation, and image classification.
With the release of version 8.1.0 on January 10, 2024, a pivotal feature was introduced: Oriented Bounding Box (OBB) models.
Unlike traditional object detection models, where bounding boxes are aligned with the image axes, OBB models incorporate an additional angle parameter, allowing for more precise localization of objects, particularly those with irregular orientations. 
This advancement holds significant potential for applications in remote sensing, where accurately detecting elongated objects is critical.

\textcolor{dg}{
Given their elongated and homogeneous nature, roof edges are ideal candidates for OBB detection. 
Additionally, the vectorized output simplifies the conversion of detected roof edges into structured vector formats, facilitating the generation of complete vector representations through subsequent polygonization.
}

\textcolor{mei}{Specifically, our model is built on the pre-trained YOLOv8 OBB model provided by Ultralytics, using its default hyperparameters, with model weights ``yolov8l-obb.pt''. To adapt our training data to OBB, we generated an approximate detected bounding box for each edge. The trained model achieved a precision of 0.99 and recall of 0.96 in edge detection. The mAP50 is 0.98 and mAP50-95 is 0.77.}

 \begin{comment}
\subsubsection{SAM}

\end{comment}
\subsection{Polygonization}\label{sec:poly}
%\dimi{???Qipeng, please don't write post-processing here! It always has the touch of smth optional. We call this section Polygonization.???}
\subsubsection{Edge Complementation}\label{sec:edges}

\begin{figure}[h!]
\begin{center}
		\includegraphics[width=1.0\columnwidth]{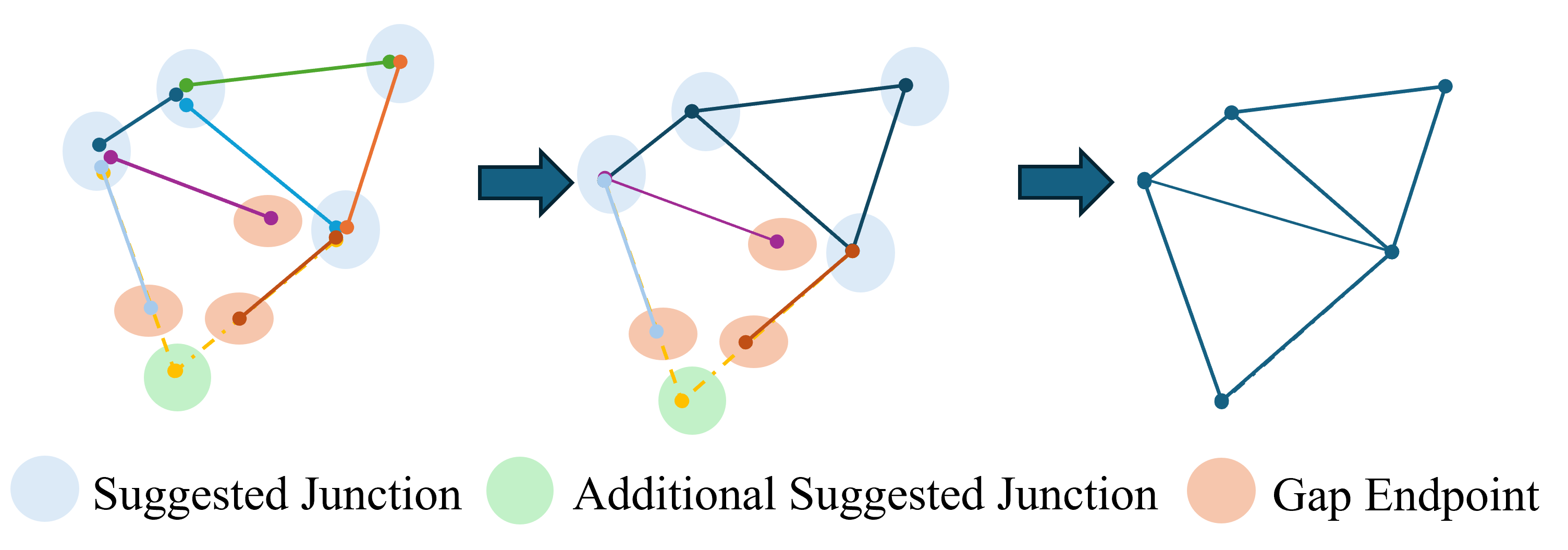}
	\caption{The process of edge complementary. }
\label{fig:line_seg}
\end{center}
\end{figure}

From the YOLOv8 output, we obtain attributes about the length and direction of edges and can infer their endpoints coordinates. 
However, it is worth noting that these edges often have gaps of varying sizes.
Therefore, we develop a polygonization procedure based on geometric rules to complement the edges, aiming to generate complete and closed polygons (see Figure~\ref{fig:line_seg}).

Firstly, we apply DBSCAN to the endpoints of edge.
It is a clustering algorithm for grouping densely distributed points, which is common for downsampling point clouds and suppressing noise \citep{sander1998density}. 
\textcolor{mei}{More than a quarter-of-century old, this method is still widely applied because of its flexibility regarding the number of clusters, robustness against noise, and high processing velocity. In our situation, where the number of clusters is uncertain and the intersections are generated through mutual constraints between different edges, this method is particularly suitable.
We differentiate between} two types of clusters: \textbf{Junction clusters} contain two or more endpoints, and we consider these points to form a junction of two or more edges. 
Contrarily, \textbf{Gap clusters} contain only one endpoint, this means that the YOLOv8 detection box cannot fully cover this edge, hence, we need to extend it appropriately to find the possibility of a closed polygon. 
\textcolor{mei}{Junction clusters resp. gap clusters are supposed to represent two types of problems while dealing with images: Noise and occlusions.}

\begin{comment}
, employed, for example, by \cite{bulatov2017chain} to simplify the vector structure of the road network and to recognize junctions or dead ends.
Inspired by them, in our case, it is used to cluster the endpoints of edge based on their spatial proximity to identify the potential junction points. 

More specifically, there are two type of clusters:
\begin{itemize}
\setlength\itemsep{0em}\setlength\parskip{0em}\setlength\topsep{0em}\setlength\partopsep{0em}\setlength\parsep{0em} 
\item{\textbf{Junction cluster} contains two or more endpoints, we consider these points to form a junction of two or more edges. } 
\item{\textbf{Gap cluster} contains only one endpoint, it means that the YOLOv8 detection box cannot fully cover this edge, so we need to extend it appropriately to find the possibility of a closed polygon. } 
\end{itemize}
\end{comment}

For the junction cluster, we update it to be the center of all points in the cluster and store this point as the suggested junction.
Considering that the junction may be formed by two incompletely covered edges, we compute the intersection of two incompletely covered edges and store it as an additional suggested junction.
Completion of gap cluster is achieved based on all the two types of junctions. 
% First, we select the suggested junctions whose distance from the line where the edge with gap is located are below a certain threshold 
%\dimi{???This is a problem Dorota was mentioning: We should not write things like certain threshold. Especially because DBSCAN itself needs a threshold. Why don't we write that we apply DBSCAN to fix the most evident junctions and then apply SAM to fix the gap endpoints???}.
% Then, the endpoint on the gap is updated to the nearest selected suggested junctions.
More specially, for an endpoint $\x$ within a gap cluster, we search for the nearest suggested junction on the line represented by the corresponding edge. 
Then this edge is modified by fusing the endpoint $\x$ with the suggested junction.

\subsubsection{Roof Face Vectorization}\label{sec:faces}
\begin{figure}[h!]
\begin{center}
		\includegraphics[width=1.0\columnwidth]{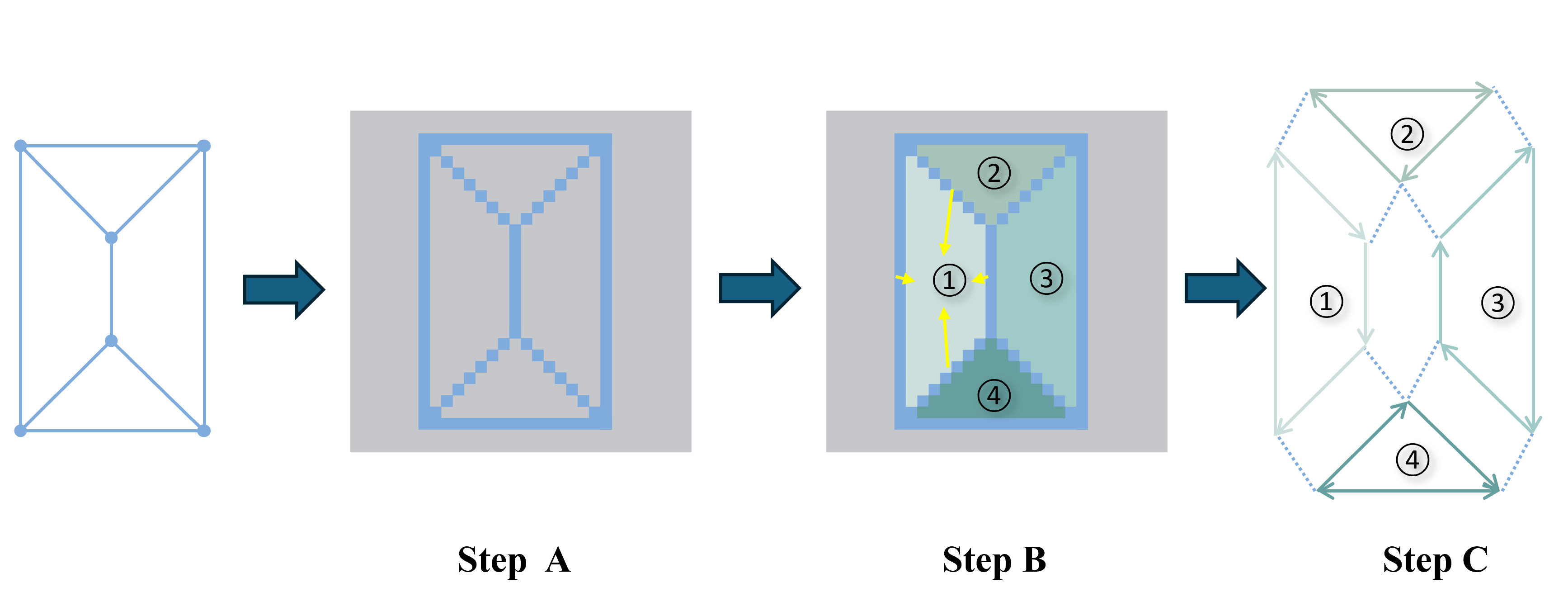}
	\caption{The process of roof face vectorization. Step A: rasterization of edges; Step B: Connected component forming; Step C: Vertices collection.}
\label{fig:point_face}
\end{center}
\end{figure}

After refining the edges, determining which points and line segment elements constitute each roof face is the final step in roof vectorization. 
To match the vertices with faces, we operate in the raster domain. 
Specifically, we first convert the edges into raster form, as shown in Step A of Figure~\ref{fig:point_face}. 
\dimi{A standard Bresenham  algorithm \textcolor{mei}{, dating back to 1965,} can be applied here.}
It produces a binary image in which the pixels occupied by edges are assigned the value 1 and the unoccupied pixels are assigned the value 0. 

Then, through connectivity checks (1-connectivity, which considers the four-neighborhood of a pixel), we partition the pixels in the raster space (see Step B of Figure~\ref{fig:point_face}). 
For a building, the partitioning result is $N+1$ components, representing the various roof faces ($N$) and the background ($1$).

Next, we label the edges from $1$ to $M$. 
By collecting the incident edges for each component and knowing the incidence relations between edges and endpoints, we can index the vertices of the face and represent it in vector form (see Step C of Figure~\ref{fig:point_face}).

\section{RESULTS}\label{sec:results}
\subsection{Datasets}\label{sec:data}
We used a building-image paired dataset created by \citet{ren2021intuitive}.
%\dimi{Name of the dataset? Country? Resolution? around 0.05m (calculate by car)? Division between training and test?}
The dataset, \dimi{denoted here as \textbf{SGA}, contains more than 3K samples of residential buildings from presumably different countries and the corresponding labels in raster and vector form}.
Each sample is an aerial image patch with the ground sample distance of \dimi{around 5cm containing} one roof of a single residential building and the label stores the 2D vertex positions of the roof vertices and the vertex-based topology of each face on the roof. 
We transformed the label into a rotation detection box for the edge of the roof to adapt it to our training of the YOLOv8 OBB model.
Then, the SGA dataset is divided into train-, val-, and test-set in a ratio of 6:2:2.
In the testset, we experiment with polygonization and evaluate the results.
%\dimi{Some ??? \% of the samples were used for training and the rest for the validation.}
\par
\dimi{The second data set has been recorded by the Spookfish (now Aerometrex) corporation in a residential district \textbf{Melville} of Perth city, Western Australia. 
The dataset has already been used in the context of geo-typical representation of the building roofs for heat simulation\textcolor{mei}{, see \cite{ilehag2018classification}}.
we represent it in the same structure of patches as in the first dataset.
The roofs exhibit very complex shapes and are difficult to annotate, \dimi{partly due to the moderate resolution of slightly below 0.1m, that is, coarser than the SGA dataset}. 
This also resulted in problems during annotation, because some roof segments are not sufficiently distinguishable by the human eye. 
In order to show the robustness of the proposed method, we decided to train our model only using input of the SGA dataset. 
For the Melville dataset, we merely annotated 50 buildings to validate our model qualitatively.}

\subsection{Evaluation Strategy}\label{sec:eval}
\subsubsection{Performance Metrics}\label{sec:metrics}
For a single building, it is important to provide evaluation on both raster and vector levels. 
%On the raster level, good reconstruction of rood details would mean that none of these details has been forgotten and also no over-generalization, such as assigning the whole roof to one planar segment (LOD0 or a shed roof) takes place. 
%Vector-based evaluation is equally important: A long narrow stripe, would not provoke a high deviation on the raster level, however, it does not look visually appealing. 
%\par
The most common function for raster-based assessment is the average metric on intersection over union (mIoU). 
Firstly, we calculate the IoU for each face of the roof.
The set of pixels belonging to the face $i$ in the reference ($R$) is represented as ${a_i}\in{R}$ while that belonging to the face $j$ in the prediction ($P$) that has the largest overlap with $a_i$ is ${b_j}\in{P}$. 
%Then, the union of the two sets, ${a_i}\cup{b_j}$, is the set of all pixels that belong to either $a_i$ and $b_j$, while the intersection, ${a_i}\cap{b_j}$, is the set of pixels that belong to both $a_i$ and $b_j$.
%Therefore, we can according to the spatial correspondence to obtain the $IoU$ of a single face.
We obtain the IoU of a single face $i$ and mIoU of the whole building according to

\begin{eqnarray}
  \mbox{IoU}_i = \frac{{a_i}\cap{b_j}}{{a_i}\cup{b_j}} \mbox{ and } \mbox{mIoU} = \frac{1}{N} \sum_{i = 1}^{N}{\mbox{IoU}_i},\label{eq_mIoU}
\end{eqnarray} 
respectively, whereby $N$ is the number  of faces in a roof. \dimi{Note that this measure is not symmetric: we can have many false roof segment hypothesis outside of $P$ without causing any harm to mIoU in \eqref{eq_mIoU}. This is why we additionally considered}
%And the $mIoU$ is the average of the $IoU$ of $N$ faces on a roof.
%$$ mIoU = \frac{1}{N} \sum_{i = 1}^{N}{IoU_i} $$
%???? Equation here please ????
%Besides, 
the overall IoU (ovIoU) computed roof-level according to $P$ and $R$, which are the total set of pixels belonging to the roof in the reference and prediction, respectively:
\begin{eqnarray}
    \mbox{ovIoU} = \frac{{P}\cap{R}}{{P}\cup{R}}.\label{eq_ovIoU}
\end{eqnarray} 
%??? Equation here please???
Both measures have the advantage that the number of segments in reference and prediction are not supposed to coincide. 

On the vector level, for every vertex $\x\in \partial P$ of the prediction polygon $\partial P$ (we omit $\partial$ in what follows), one must compute the closest point $\y \in R$ and, in the next step, vice versa. The corresponding distance 
\begin{eqnarray}
    d_{P\rightarrow R}(\x) = \min_{\y} \mbox{dist}(\x, \y)
\end{eqnarray}

\noindent must be aggregated -- somehow -- over all $\x \in P$. The symmetric Hausdorff distance presupposes taking the maximum 
 \begin{eqnarray}
    d_H = \max\left (\max_{\x \in P} d_{P\rightarrow R}(\x), \max_{\y \in R} d_{R\rightarrow P}(\y)\right)
\end{eqnarray}
while the PolyS metric of \citet{avbelj2014metric} presupposes computation of the RMSE values
\begin{eqnarray}
    % d_P = \frac{1}{2}\left(\sum_{\x \in P} d^2_{P\rightarrow R}(\x)\right)^{\frac{1}{2}}
    % +\frac{1}{2}\left(\sum_{\y \in R} d^2_{R\rightarrow P}(\y)\right)^{\frac{1}{2}}.
    d_P &=& \frac{1}{2}\left(\sum_{\x \in P} d^2_{P\rightarrow R}(\x)\right)^{\frac{1}{2}} \nonumber \\
    && + \frac{1}{2}\left(\sum_{\y \in R} d^2_{R\rightarrow P}(\y)\right)^{\frac{1}{2}}.
\end{eqnarray}
Both metrics have often been applied to assess the quality of reconstruction. 
Instead of averaging the errors in $d_P$, we take the maximum of both values. 
Since we wish to have the value 1 to correspond to a good reconstruction and 0 to a bad reconstruction, we norm these metrics: $q_{.} =  1-d_{.}/d_{\max}$, thus making the uniform with \eqref{eq_mIoU} and \eqref{eq_ovIoU}. Hereby, $q$ stands for quality, $\cdot$ means either Hausdorff \textcolor{mei}{($H$)} or PolyS \textcolor{mei}{($P$)} metric, and $d_{\max}$ is a scaling parameter, a diagonal of the bounding box of both polygons. 
Finally, the vector-raster metric presupposes preserving vector and matrix properties of reference data. It is given by
\begin{eqnarray}
    q_{\scriptsize{\mbox{VM}}} = \mbox{mIoU}\cdot q_H
\end{eqnarray}
and is scaled between 0 (bad on the raster \emph{or} vector level) and 1 (good on the vector \emph{and} raster level), as well.
\par
The single error metrics are given building-wise. Since our datasets consist of many buildings, we wish to assess the reconstruction accuracy by the average values of mIOU, OvIoU, $q_P$, and $q_{\mbox{\scriptsize{VM}}}$ as well as the median of $q_H$, because Hausdorff metric already takes gross errors and outliers into account. 
Additionally, we will show the \dimi{boxplots} of \textcolor{mei}{all metrics}.
\par
\subsubsection{SAM as Competing Approach}\label{sec:SAM}
%\begin{itemize}
%    \item Short description of this method
%    \item How we use seed points to obtain regions
%    \item We only will use it to compare with our method!
%\end{itemize}
While our approach heavily relies on roof edges, we wonder to what extent modern foundational models are suitable to retrieve roof faces directly. 
To this end, we use the prediction of the Segment Anything Model (SAM) as the baseline of our polygonization at raster level.
SAM is a foundation model for image segmentation released by Meta \citep{kirillov2023segment}. 
It consists of three main modules:
%\begin{itemize}
 %   \item 
 1) Image encoder: composed of MAE pre-trained Vision Transformer (ViT), which maps the image to be segmented into an image feature space;
 %   \item 
 2)   Prompt encoder: responsible for mapping the input prompt to a prompt feature space;
  %  \item 
3)    Mask decoder: integrates the embedding output by the Image encoder and Prompt encoder, then decodes the final segmentation mask from this embedding's feature.
%\end{itemize}
\par
Trained on a massive dataset (SA-1B, comprising 1B masks and 11M images), SAM has developed powerful generalization capabilities, enabling 
%This allows it to 
transfer to new image distributions and tasks through zero-shot learning. 
Users can interact with SAM using prompts (e.g., points, boxes, masks) to obtain relevant segmentation masks. 
In our research, we designed a prompt generation strategy to use SAM to obtain masks for each face of building roofs, which we then compared with our method's results.
The prompts of the SAM can be a set of foreground/background points. 
Therefore, we designed a prompt generation strategy. 
First, we sample each face $f$ of the roof based on the ground truth, \textcolor{mei}{which plotted from the roof face vector data in the dataset}, to obtain a set of prompt points.
%\textcolor{mei}{This is an application of prompt engineering, and in future work we would improve the method to enable prompt generation on unreferenced inputs.???}
\begin{figure}[ht!]
\begin{center}
		\includegraphics[width=1\columnwidth]{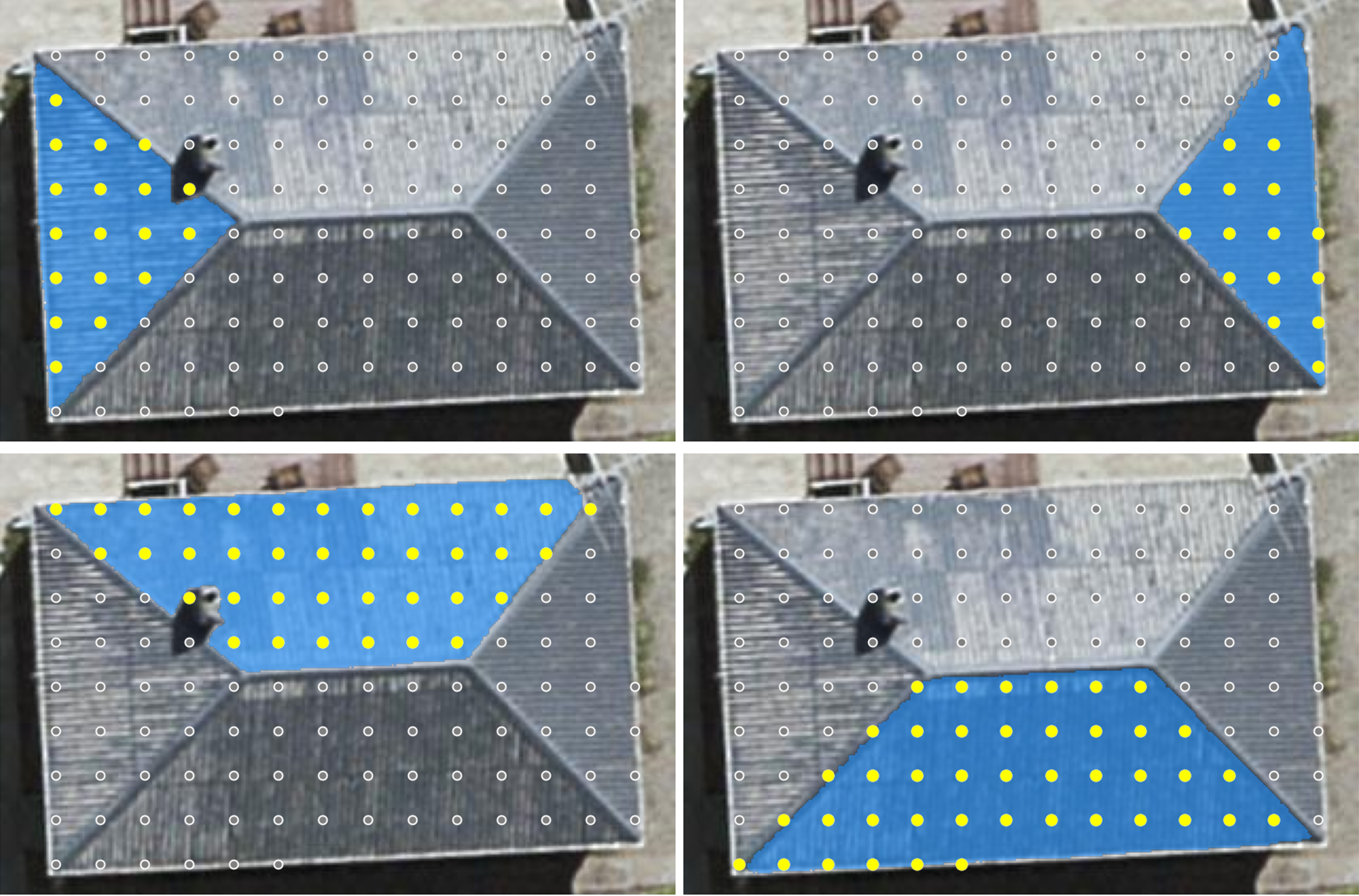}
	\caption{Example of prompt generation strategy. (Foreground/background prompts are denoted by yellow and gray points, respectively, while the blue mask denotes the prediction of SAM.)}
\label{fig:sam_example}
\end{center}
\end{figure}

Then, we divide the prompt set into two subsets: 
The subset belonging to $f$ is used as foreground prompt points and the remaining as background prompt points, as shown in Figure \ref{fig:sam_example}.
\textcolor{mei}{Using these prompt points, SAM is able to segment each individual face and delineate the boundary. 
As a result, we obtain the prediction results of the foundational model under strongly directional prompts.}
%Then, we can obtain the prompt-based prediction of SAM for each face.

% \begin{figure}[h!]
%     \centering
    
%     \begin{subfigure}[b]{0.23\textwidth}
%         \centering
%         \includegraphics[width=\textwidth]{figures/sam/sam_1.png}
%         \caption{Left Face}
%         \label{fig:sam_l}
%     \end{subfigure}
%     \hfill
%     \begin{subfigure}[b]{0.23\textwidth}
%         \centering
%         \includegraphics[width=\textwidth]{figures/sam/sam_2.png}
%         \caption{Top Face}
%         \label{fig:sam_t}
%     \end{subfigure}
    
%     \vspace{0.5cm}
    
%     \begin{subfigure}[b]{0.23\textwidth}
%         \centering
%         \includegraphics[width=\textwidth]{figures/sam/sam_3.png}
%         \caption{Right Face}
%         \label{fig:sam_r}
%     \end{subfigure}
%     \hfill
%     \begin{subfigure}[b]{0.23\textwidth}
%         \centering
%         \includegraphics[width=\textwidth]{figures/sam/sam_4.png}
%         \caption{Bottom Face}
%         \label{fig:sam_b}
%     \end{subfigure}
    
%     \caption{\dimi{Subfigures not really sensible here}
%     Example of Prompt Generation Strategy. \dimi{Foreground background prompt are denoted by green and gray points, respectively, while the blue mask denotes the prediction of SAM.}}
%     \label{fig:sam_example}
% \end{figure}

\subsection{Findings}\label{sec:findings}

\subsubsection{Quantitative Evaluation}\label{sec:quantitative}

\begin{figure}[htbp]
    \centering
    \begin{subfigure}{0.184\textwidth}
        \centering
        \includegraphics[width=\textwidth]{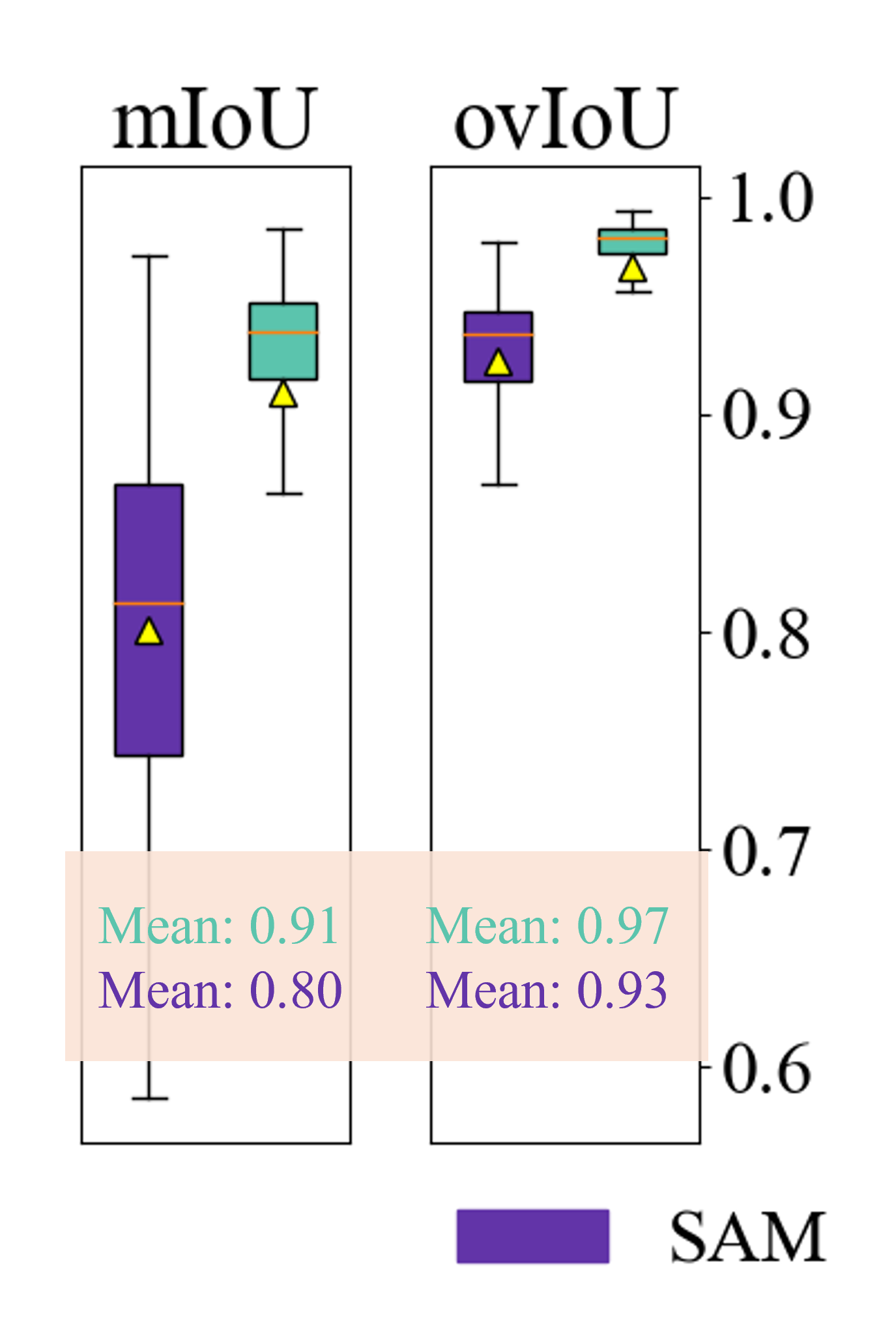}
        \caption{Raster level}
        \label{fig:raster}
    \end{subfigure}
    \hfill
    \begin{subfigure}{0.282\textwidth}
        \centering
        \includegraphics[width=\textwidth]{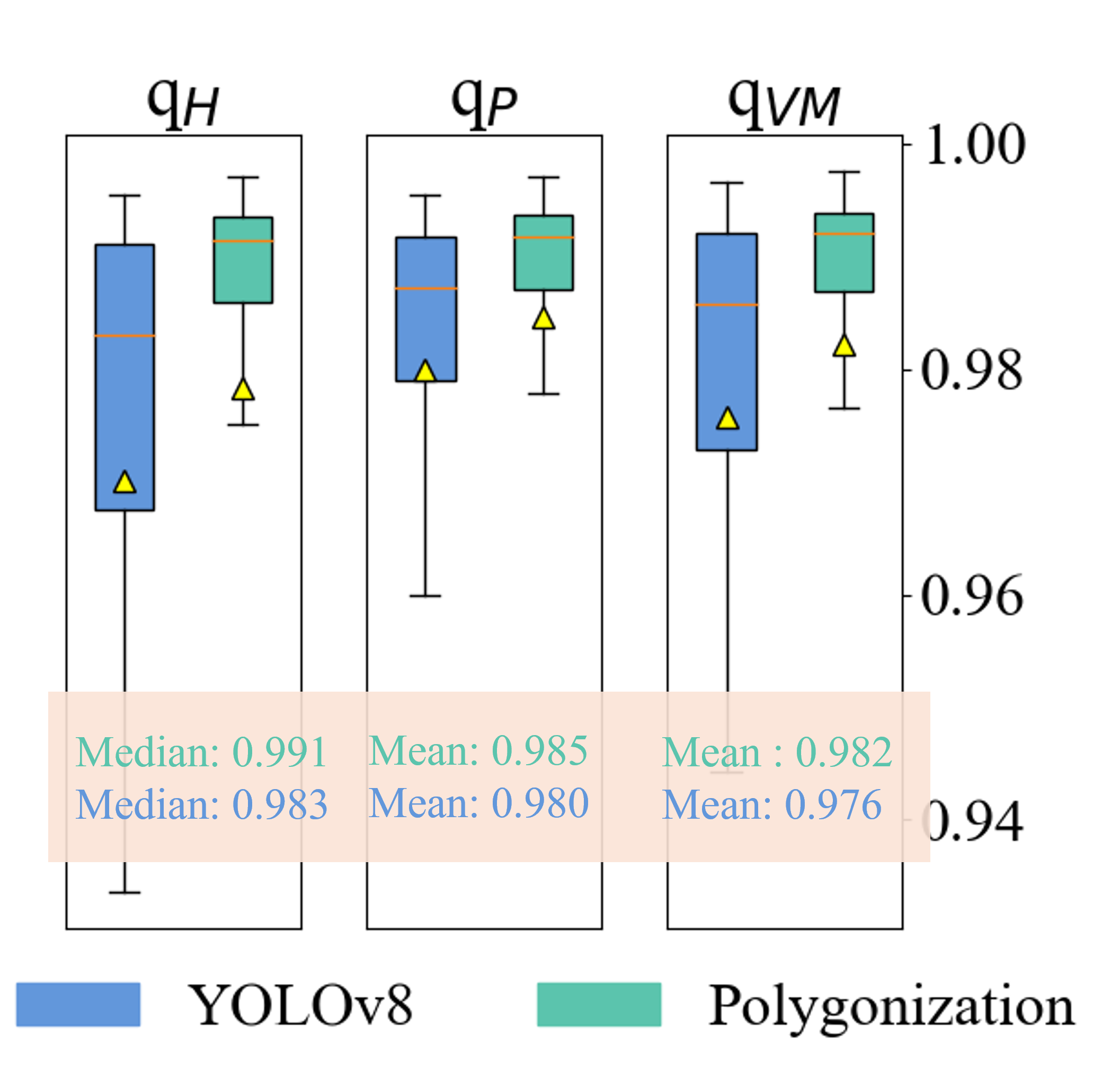}
        \caption{Vector level}
        \label{fig:vector}
    \end{subfigure}
    \caption{Quantitative evaluation of the SGA dataset.}
    \label{fig:quantitative_evaluation}
\end{figure}

Figure~\ref{fig:raster} is the quantitative evaluation on the raster level. 
We compare our method and the prediction of SAM, with outliers removed for a clearer view of the majority.
%The first box plot presents the IoU distribution of each face for both methods.
First of all, it is evident that our method outperforms SAM in terms of mIoU. 

The distribution of our method is heavily skewed towards higher mIoU, for most roofs falls within the range of 0.85 to 1, with a mean value of 0.91, indicating excellent segmentation performance.
 In contrast, SAM's distribution is more spread out and falls within the range of 0.6 to 1.
 And its mean value is 0.8.

 In terms of ovIoU, the difference between SAM and our method is relatively small.
However, our method maintains ovIoU within a narrower range of 0.95 to 0.99, while SAM achieves a range of 0.85 to 0.98.
This indicates that our approach provides more consistent coverage of buildings.

Figure~\ref{fig:vector} presents the quantitative evaluation at the vector level, with outliers removed for a clearer view of the majority.
Across the three metrics, the outputs of YOLOv8 closely approximate the ground truth, with most samples having a $q_H$ between 0.97 to 0.99, a $q_P$ between 0.98 to 0.99 and a $q_{\scriptsize{\mbox{VM}}}$ between 0.97 to 0.99. 
After polygonization, the sample distribution becomes more concentrated, indicating further optimization of YOLOv8.
This improvement is also reflected in the considerably higher median lines across all metrics compared to YOLOv8.
Consequently, our method demonstrates better consistency.

The $q_H$, which is particularly sensitive to outliers, is of special interest to us.
It reveals that YOLOv8's lower whiskers extend further, indicating the presence of some gross errors during evaluation.
However, it's important to note that these outliers are effectively suppressed by our polygonization process, as can be seen from the increase in median value from 0.983 to 0.991, demonstrating the robustness of our method.
The PolyS metric $q_P$ reveals that YOLOv8 effectively identifies most roof edges, but struggles with completeness, resulting in relatively low values of $q_p$. 
Our polygonization method, however, refines these roof edges, significantly enhancing geometric similarity to the reference object. 
Finally, The $q_{VM}$ metric, which combines raster-level and vector-level evaluations, shows minimal divergence from $q_P$ and $q_H$ of our method. 
This indicates excellent performance in both geometric and raster accuracy for most samples, yielding a reliable and robust vector representation of the roof structure.

Overall, at the vector level, YOLOv8's output is satisfactory, and our polygonization further enhances the results. 
While effectively representing the roofs in the vector form, our method demonstrates strong robustness and consistency.

\subsubsection{Qualitative Evaluation}\label{sec:qualitative}
%\vspace{2mm}
\begin{figure*}[htp!]
    \centering
    \begin{subfigure}{0.48\textwidth}
        \fbox{\includegraphics[width=\linewidth]{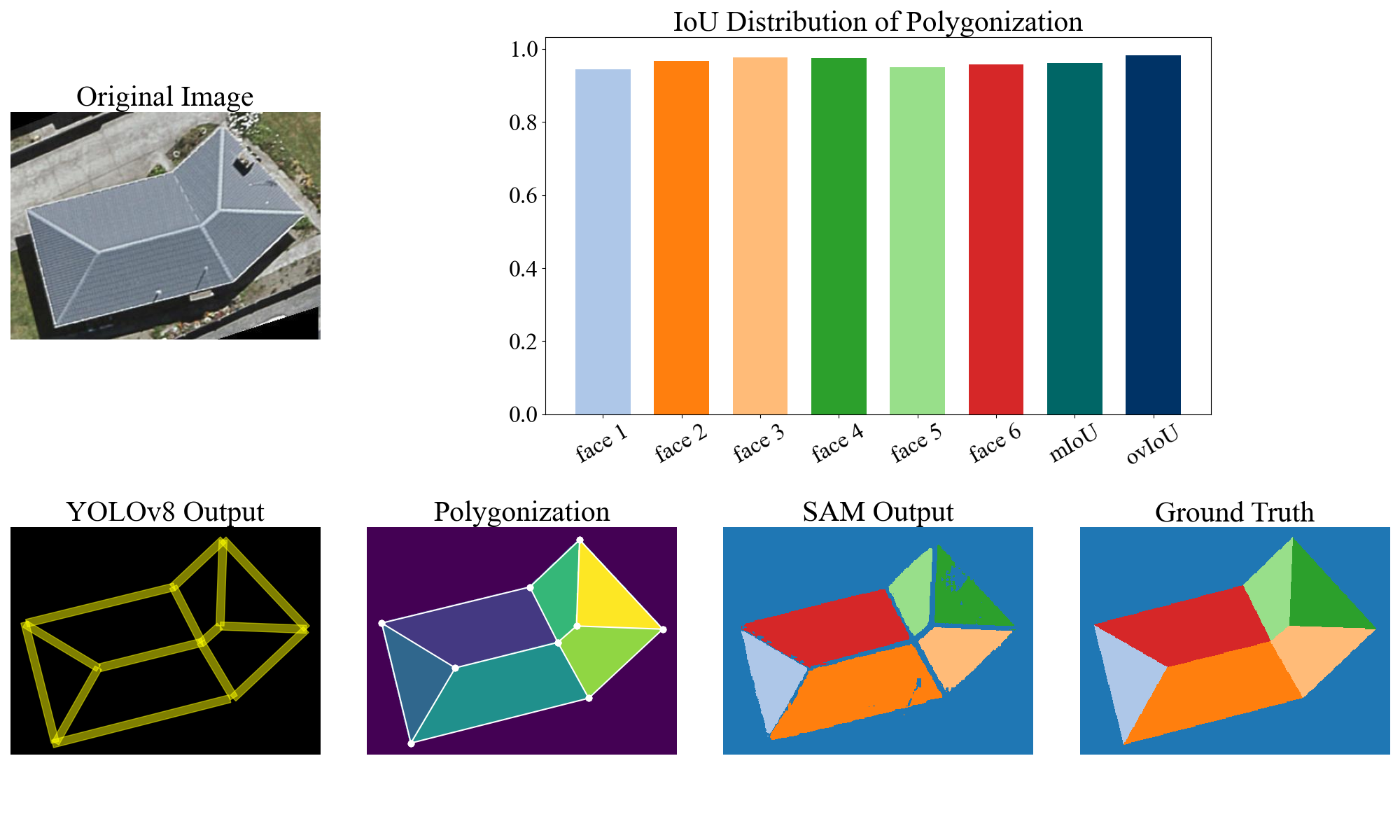}}
        \caption{Simple case}
    \end{subfigure}
    \hspace{0.01\textwidth} 
    \begin{subfigure}{0.48\textwidth}
        \fbox{\includegraphics[width=\linewidth]{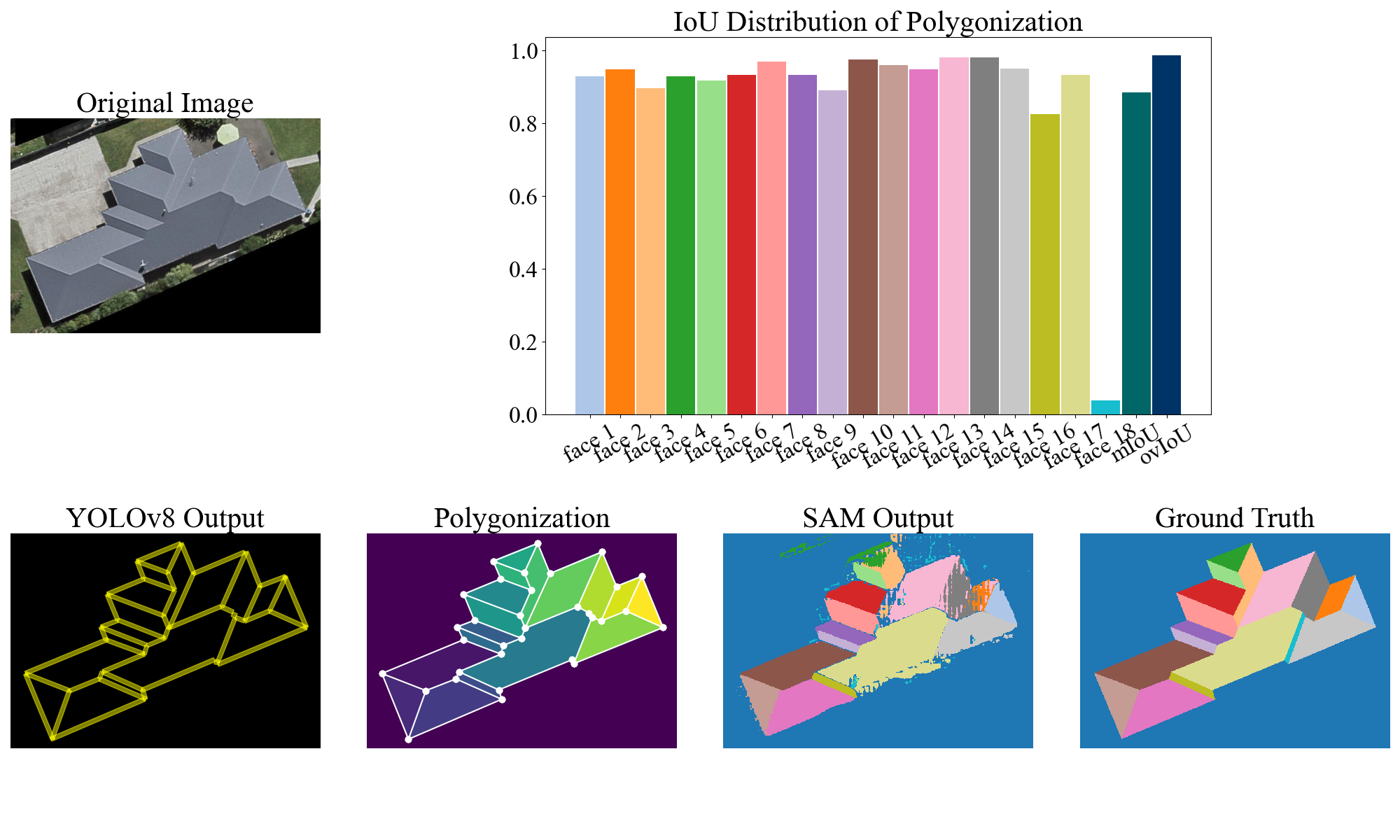}}
        \caption{Complex case}
    \end{subfigure}
    
    \vspace{0.01\textwidth}
    \begin{subfigure}{0.48\textwidth}
        \fbox{\includegraphics[width=\linewidth]{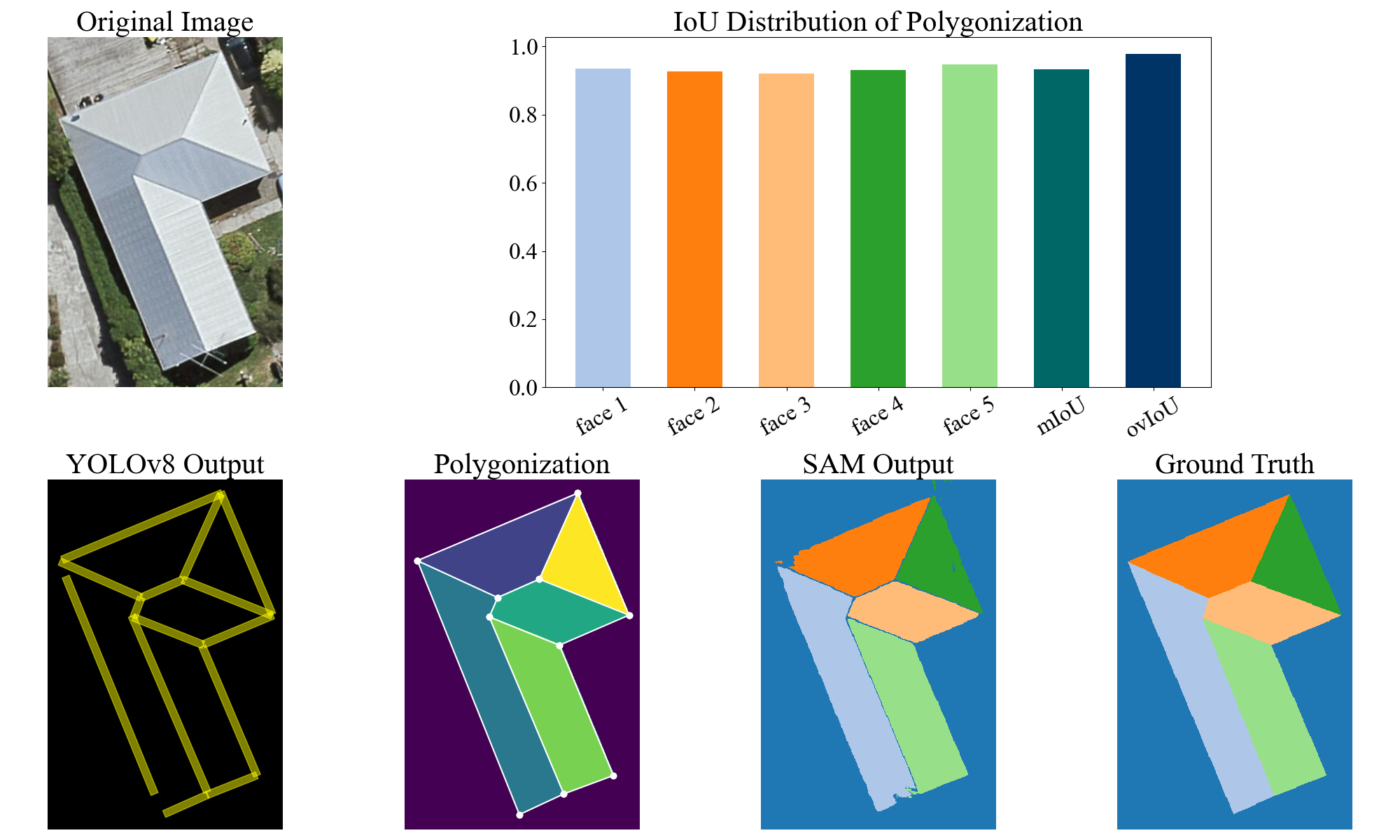}}
        \caption{One edge gap}
    \end{subfigure}
    \hspace{0.01\textwidth}
    \begin{subfigure}{0.48\textwidth}
        \fbox{\includegraphics[width=\linewidth]{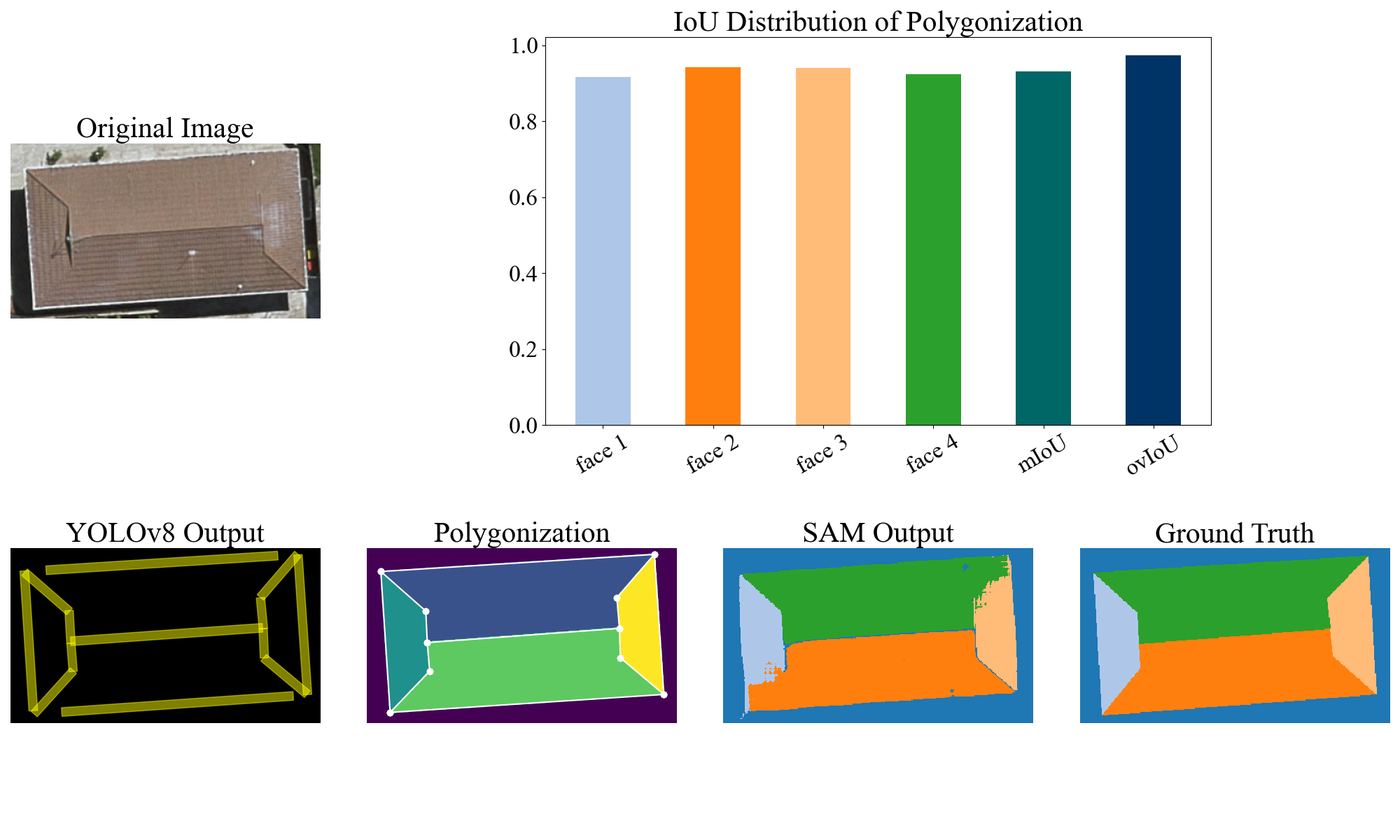}}
        \caption{Two edge gaps}
    \end{subfigure}
    
    \vspace{0.01\textwidth}
    \begin{subfigure}{0.48\textwidth}
        \fbox{\includegraphics[width=\linewidth]{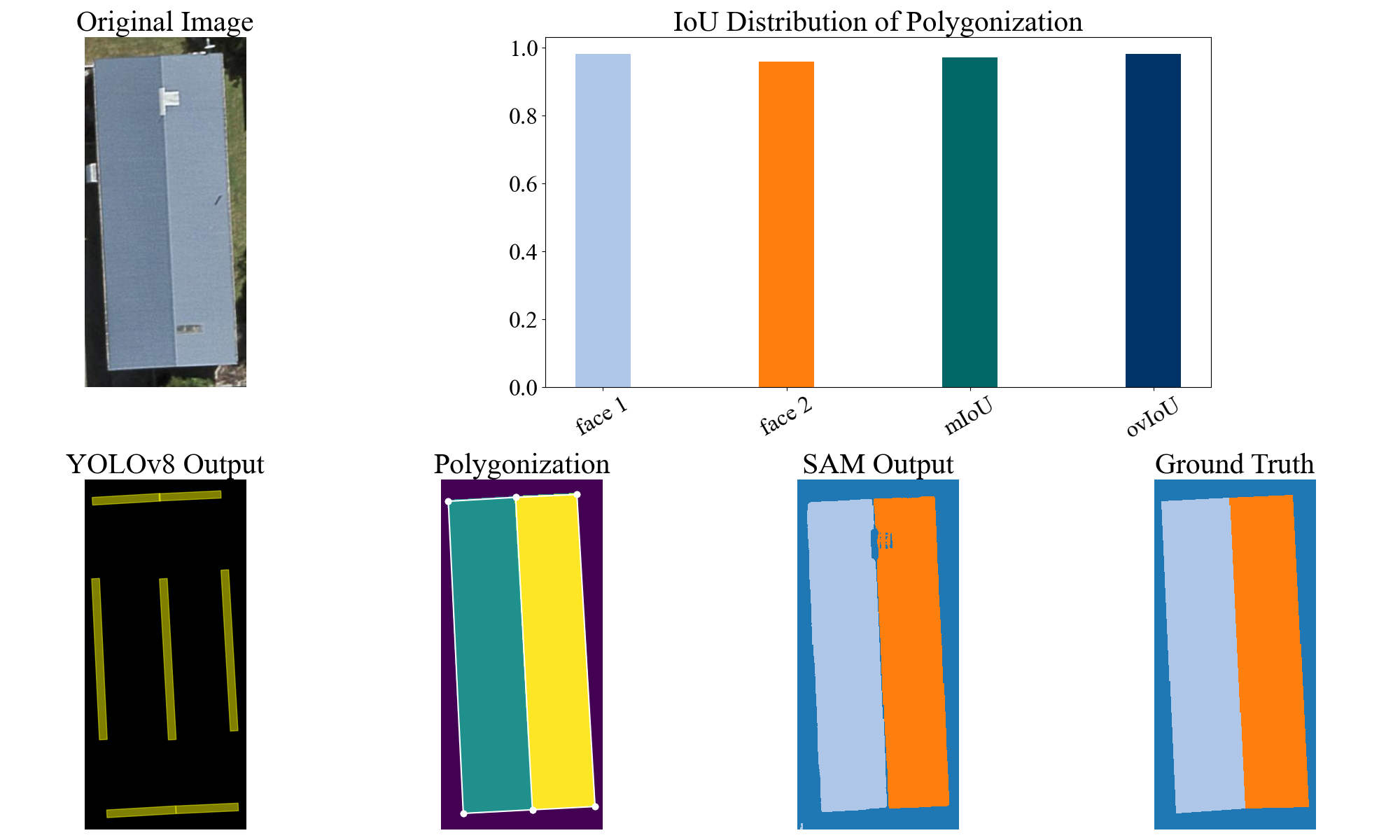}}
        \caption{Multi edge gaps}
    \end{subfigure}
    \hspace{0.01\textwidth}
    \begin{subfigure}{0.48\textwidth}
        \fbox{\includegraphics[width=\linewidth]{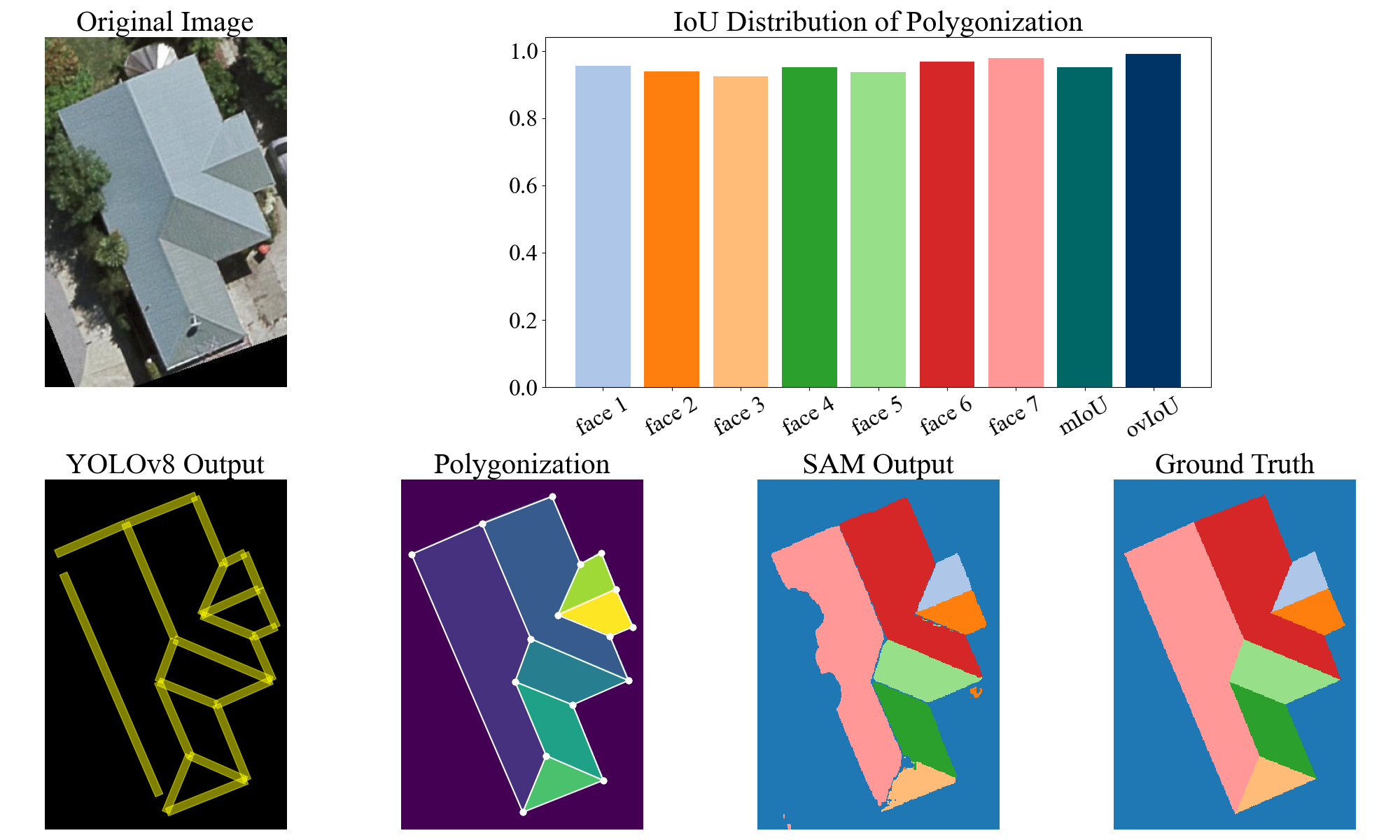}}
        \caption{Tree cover}
    \end{subfigure}

    \vspace{0.01\textwidth}
    \begin{subfigure}{0.48\textwidth}
        \fbox{\includegraphics[width=\linewidth]{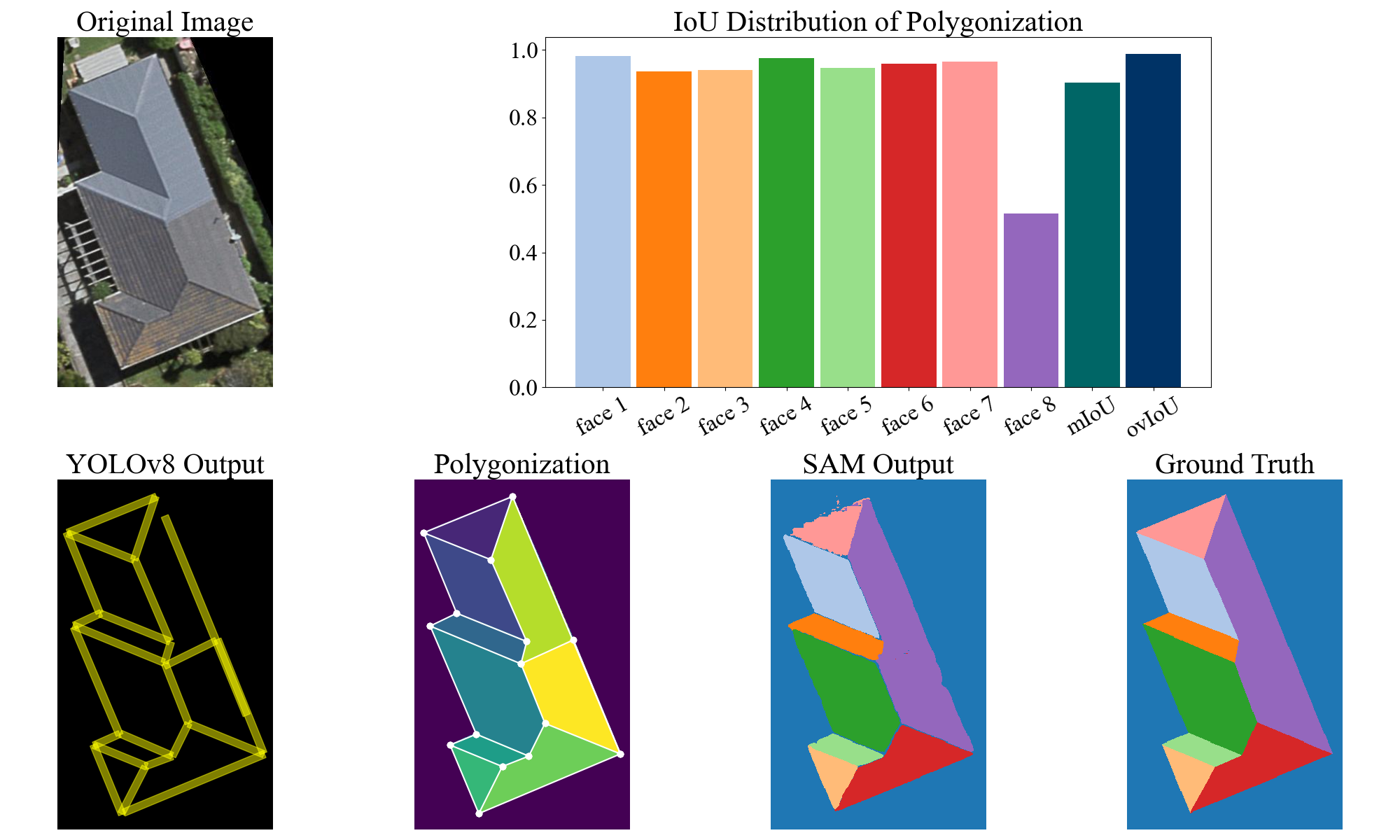}}
        \caption{Detect error}
    \end{subfigure}
    \hspace{0.01\textwidth}
    \begin{subfigure}{0.48\textwidth}
        \fbox{\includegraphics[width=\linewidth]{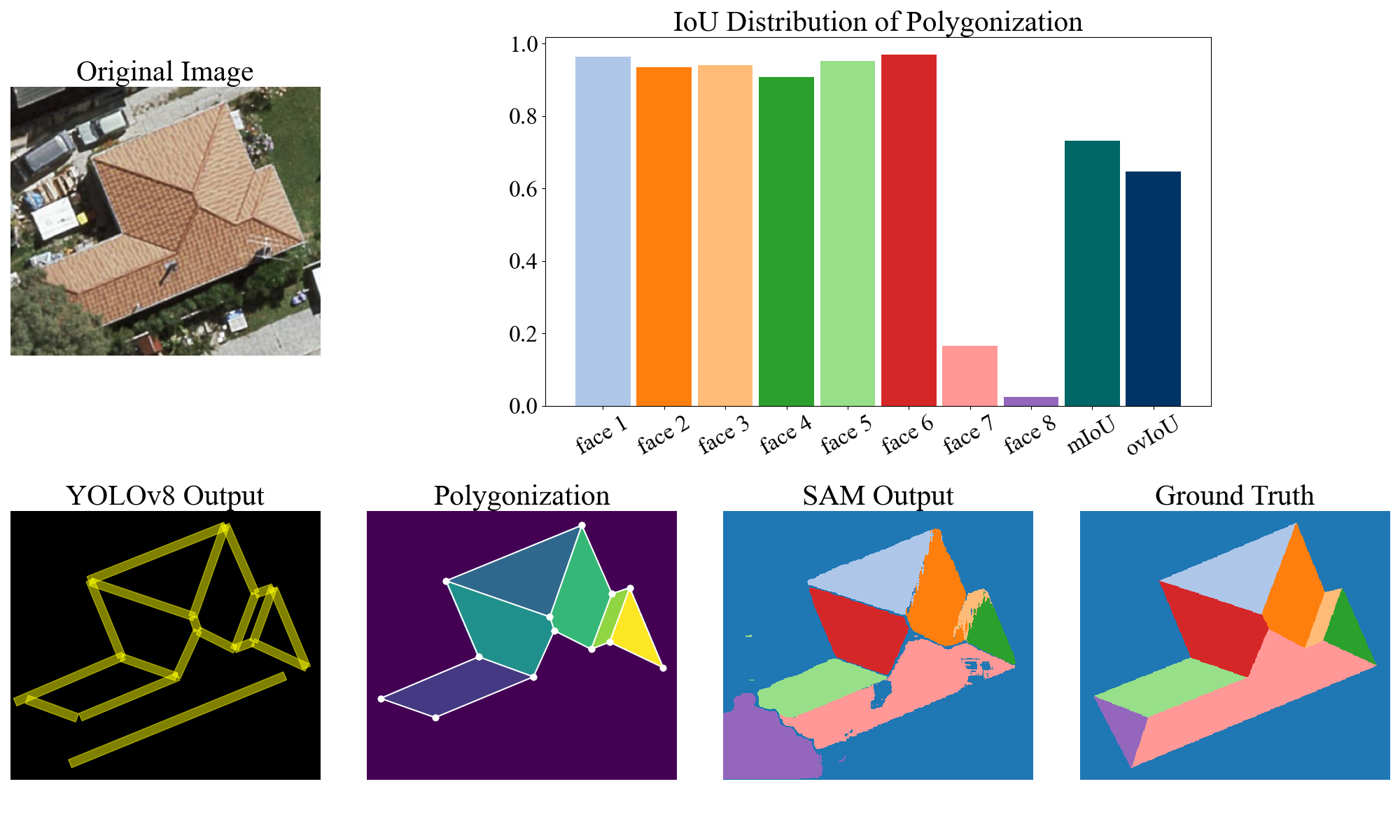}}
        \caption{Edge loss}
    \end{subfigure}
    \caption{Qualitative evaluation samples (SGA). }
    \label{fig:qualitative}
\end{figure*}

Figure~\ref{fig:qualitative} presents a qualitative analysis of eight representative samples from the SGA dataset.
For each sample, we display its aerial image, the detection results from YOLOv8 OBB (with detected bounding boxes shown in yellow), the polygonization results (final representations of roof structures with vertices and edges shown in white), the SAM prediction results, and the ground truth (individual faces represented as colored areas).
The colors of the SAM outputs and the ground truth correspond directly to the colors of each face in the IoU distribution. 
Additionally, the mIoU and ovIoU values are provided.

Sample (a) demonstrates a simple case.
The accurate prediction of YOLOv8 enables our polygonization to complete the roof vectorization in the first step. 
In the SAM results, each face is also well predicted. 
Through further observation of the IoU for each face, the polygonization results are satisfactory.
Sample (b) is a complex roof case.
Despite the complexity of the roof structure, most of the edges are well recognized.
The lack of prediction of face edges18 leads to its low IoU.
However, for most faces, the results are satisfactory, and the ovIoU is also at a high level.

Sample (c) to (h) have varying degrees of edge gaps.
The gap in Sample (c) is located on the longest edge of the roof, at the lower left of the building.
And the gaps in Sample (d) are present on the upper and lower outer contour edges.
Our polygonization is able to effectively handle these situations.
The edge implementation approach matches the gap endpoints to appropriate potential junctions, thus achieving a complete roof topology.
As a result, each face obtains excellent IoU.
% Examples 3 to 8 all have varying lengths of edge gaps. The gap in Example 3 is located on the longest edge of the roof, at the lower left of the building. For this situation, our Edge Complementation matches it to appropriate potential nodes, thus achieving a complete roof topology. As a result, each face shows excellent IoU. In Example 4, both the upper and lower outer contour edges have considerable gaps, which our method can effectively handle. 

Sample (e) demonstrates the effectiveness of additional junctions in our edge completing approach.
The presence of additional junctions allows its contour edges to be well completed.
From Samples (c) to (e), we can see that gaps mainly exist in longer edges of building roofs, and are more common in outer contour lines. 
Possible reasons include: compared to shorter edges, long edges have fewer samples; hence, YOLOv8 may not learn them sufficiently; additionally, the prediction tends to cover areas with higher confidence, which may sacrifice some areas that could be edges. 
Nevertheless, our polygonization compensates well for this weakness, using geometric prior knowledge to connect these edges with appropriate junctions.

Sample (f) demonstrates an advantage of our method: on buildings covered by trees, edge as a geometric primitive have a better spatial continuity, thus overcoming the effects of visual occlusion.
The lower left corner edge in this example is occasionally obscured by trees, which may present a challenge for algorithms that use corners as geometric primitives.
The SAM results also show that prompt-based Segmentation cannot cover this face well.
However, from the vector prediction of YOLOv8, this face is still well reconstructed with the assistance of our polygonization.

Sample (g) illustrates a weakness of our method, namely if the roof is composed of textures from two different materials, this texture may create a false edge.
Then, our method might incorrectly divide this face, causing prediction errors. 
Surprisingly, SAM was able to achieve a better result. 
Finally, Sample (h) represents a situation our method cannot handle. 
Here, one face is completely occluded by trees, resulting in up to three edges not being recognized, thus causing the adjacent face to be missed. 
Certainly, this also poses a challenge to SAM, which has prompts situated on trees and results as mask of tree.
For such cases, we can only seek help from other data sources to achieve roof reconstruction.

Overall, our method can effectively handle various roof scenarios with clear geometric relationships, and also demonstrates well robustness for partially obscured edges.

\begin{figure}[htp!]
\begin{center}
		\includegraphics[width=1.0\columnwidth]{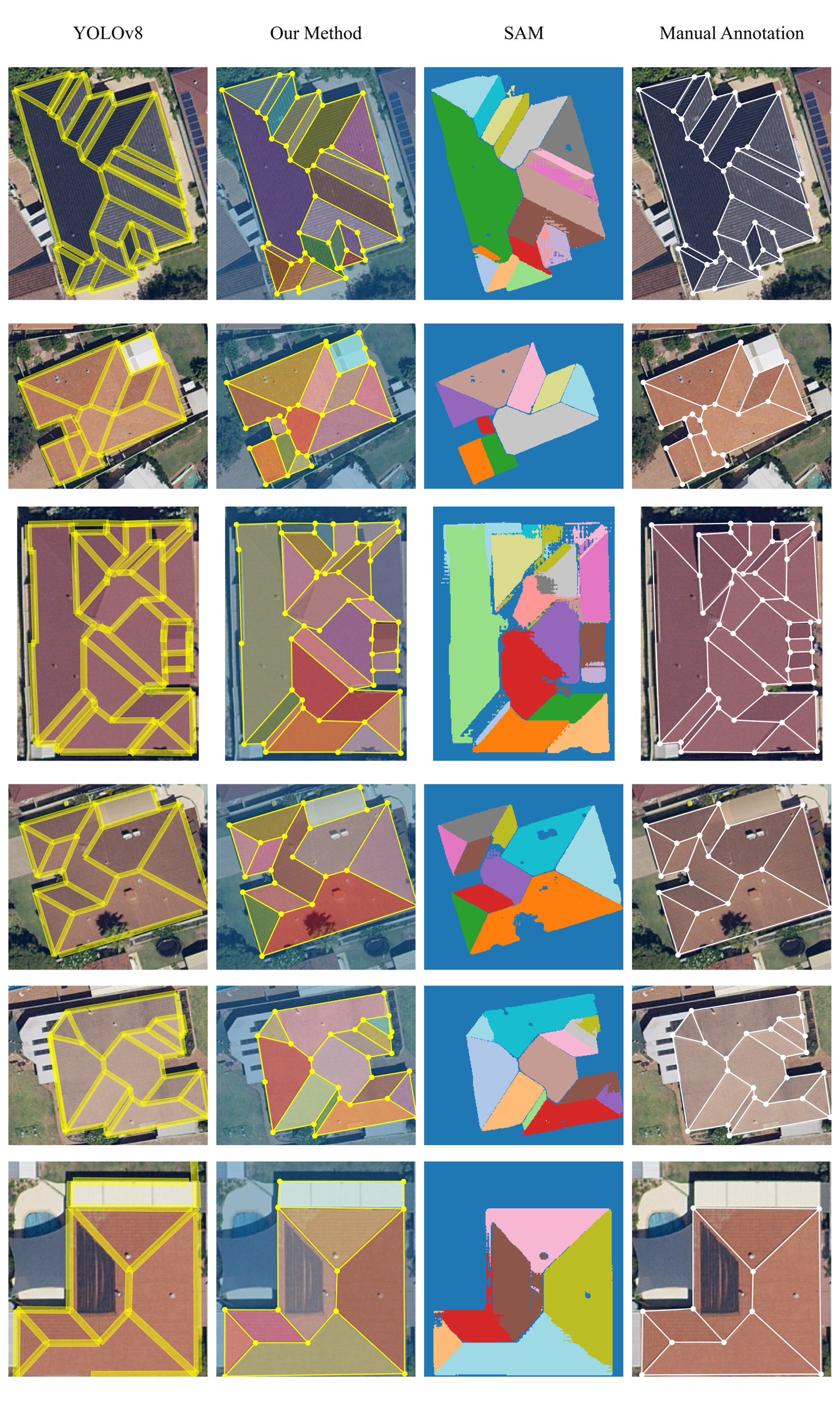}
	\caption{Qualitative evaluation samples (Melville). From left to right: Input patch mit YOLOv8 lines, Polygonization output, SAM-output, ground truth delineation.}
\label{fig:melville_qualitative}
\end{center}
\end{figure}

Turning our attention to the dataset Melville, we present in Figure~\ref{fig:melville_qualitative} a comparison between the roof structure vector representations generated by our method and the manual annotation. 
Lines represent detected roof edges, while points indicate the corners of these edges. 
The top row, for example, represents a very complex building, all edges of which were detected by YOLOv8, followed by one spurious fusion caused by our polygonization method. 
SAM, however, could not detect some narrow segments. 
The bottom row represents a building quite typical for the SGA dataset. 
It is not very complex, but one solar panel on the roof makes the SAM algorithm struggle. 
Our method reconstructs this building completely and correctly.
Although the Melville dataset was not involved in our model training, our method's output shows a generally consistent pattern with manual annotations. 
The main shortcomings lie in the accuracy of some roof details — our method tends to ignore or simplify narrower and irregular faces. 
Such simplification may have severe repercussions in the case of 3D reconstruction of roofs; 
nevertheless, these results demonstrate our method's potential in handling complex roof vectorization and significantly reducing labor costs. 

%\vspace{-2mm}
\section{CONCLUSION}\label{sec:conclusion}
%\vspace{-2mm}

We have developed a novel method for roof detail extraction and vectorization based on remote sensing images.
%Unlike previous methods based on geometric primitives, which predominantly focus on corner points
In particular, our approach leverages edges as the primary elements for reconstruction, facilitating the determination of roof corners and faces through geometric relationships. 
The strength of this approach lies in its robustness in edge detection, enabling the reconstruction of accurate structures even when the roof is partially occluded and minimizing the impact of noise. 
This overcomes the limitations of corner-based vector structure reconstruction.

Specifically, roof edge extraction is achieved using the YOLOv8 OBB model. 
We have innovatively adapted this model, typically employed for rotated object detection, to the task of roof edge detection. 
Its vectorized output is particularly well-suited for edge extraction, simplifying the process of generating vector representations of roofs.

We conducted experiments on two datasets: the SGA dataset created by \citet{ren2021intuitive}, to the extent we could observe, without topological or geometric errors, and the Melville dataset that we annotated.
The former was also used to train our YOLOv8 OBB model and  evaluated our method at both raster and vector levels.
At the raster level, we used the SAM as the baseline, which is widely regarded as the state-of-the-art foundational model for segmentation. 
\begin{comment}
    Our model performed excellently at the raster and vector level, with most samples maintaining a mIoU between 0.85 and 1 for each roof, and an ovIoU close to 0.97.
 In comparison, SAM's performance fluctuated more.
 At the vector level, we use Hausdorff distance, the PolyS metric of \cite{avbelj2014metric} while our own $q_{VM}$ metric (integrates the raster and vector qualities) to evaluate the YOLOv8 and the optimized results.
 The results showed that the output of YOLOv8 already closely approximated the reference, with significant improvements after our polygonization approach.
 Furthermore, we successfully achieved roof structure vectorization.
\end{comment}
Our model not only performed excellently at the raster and vector level, but also produced quite stable results with narrow quantile ranges, including for the quite outlier-sensitive Hausdorff metric.
 In comparison, SAM's performance fluctuated more. 
 The main reason for this is that the edges are real, detectable observations while the roof segments SAM relies on can suffer from occlusions by trees, color changes, roof objects, and so on.
\par

We also conducted qualitative evaluations on both the SGA and Melville datasets, showcasing representative results.
Overall, our method can effectively handle different roof structures and eliminate edge gaps in YOLOv8.
Even on the Melville model, which was not involved in training and has a relatively lower resolution, the results were satisfactory.
Although the roofs in Melville present more complex structures, most roof segments could still be well recognized. 
This demonstrates our great potential in handling complex roof vectorization.

In conclusion, our findings highlight the potential of our method to effectively handle diverse roof structures, even in challenging scenarios with complex geometries. 
Moving forward, we plan to explore additional datasets and integrate our technique into urban terrain reconstruction workflows. 
\textcolor{mei}{On the one hand, it will help to explore further radiometric (detecting important installations on roofs: photovoltaic panels, solar collectors, etc.) and geometric (non-planar roof elements, such as domes, towers) aspects.
On the other hand, we} aim to conduct a more comprehensive comparative analysis with other competing methods, further establishing the robustness and versatility of our approach. 
\textcolor{mei}{Finally, in future work, we plan to incorporate 3D data to search for more precise intersections near our predicted ones to improve the vectorization.}
\vspace{-2mm}
\section*{ACKNOWLEDGEMENTS}
\vspace{-2mm}
The authors thank the China Scholarship Council (CSC) for supporting this research, Grant/Award Number: 202308080109.
We also thank the reviewers for their insightful comments. 
\vspace{-2mm}
	%\vfill
	%\vspace{-3mm}
	\bibliographystyle{apalike}
	{%\small
		\bibliography{literature}}
	%\section*{\uppercase{Acknowledgements}}

	%\section*{\uppercase{Appendix}}
	
	\noindent 
	
\end{document}